\documentclass[11pt,a4paper]{article}
\usepackage{times,latexsym}
\usepackage{url}
\usepackage[T1]{fontenc}
\usepackage{amsmath}
\usepackage{color,soul}
\usepackage{graphicx}
\usepackage{subcaption}
\usepackage{booktabs}
\usepackage{multirow}
\usepackage{caption}
\DeclareCaptionFont{xipt}{\fontsize{10}{10}\mdseries}
\usepackage[font=xipt]{caption}
\usepackage{makecell}
\usepackage{CJKutf8}
\usepackage{times}
\usepackage{latexsym}
\usepackage{hyperref}
\usepackage{breakurl}
\usepackage{bm}
\usepackage{makecell}

\usepackage[acceptedWithA]{tacl2021v1}
\usepackage{xspace,mfirstuc,tabulary}

\newif\iftaclinstructions
\taclinstructionsfalse 
\iftaclinstructions
\renewcommand{\anonsubtext}{(No author info supplied here, for consistency with
TACL-submission anonymization requirements)}
\newcommand{\instr}
\fi

\iftaclpubformat 

\else

\fi

\title{Predicting Human Translation Difficulty with Neural Machine Translation}

\author{
  Zheng Wei Lim,
  Ekaterina Vylomova,
  Charles Kemp,
  \and 
  Trevor Cohn\Thanks{Now at Google DeepMind}
  \\
  The University of Melbourne \\
  \texttt{z.lim4@student.unimelb.edu.au} \\
  \texttt{\{vylomovae,c.kemp,t.cohn\}@unimelb.edu.au}\\
}

\begin{document}

\newcommand{\att}[3]{$f^{\mathrm{#1}}_{#2,#3}$}
\newcommand{\ent}[3]{$H^{\mathrm{#1}}_{#2,#3}$}
\newcommand{\bs}[1]{\boldsymbol{#1}}
\newcommand{\nt}[1]{\mkern 1.5mu\overline{\mkern-1.5mu#1\mkern-1.5mu}\mkern 1.5mu}

\newcommand{\veryshortarrow}[1][3pt]{\mathrel{%
   \hbox{\rule[\dimexpr\fontdimen22\textfont2-.2pt\relax]{#1}{.4pt}}%
   \mkern-4mu\hbox{\usefont{U}{lasy}{m}{n}\symbol{41}}}}

\maketitle
\begin{abstract}
Human translators linger on some words and phrases more than others, and predicting this variation is a step towards explaining the underlying cognitive processes.
Using data from the CRITT Translation Process Research Database, we evaluate the extent to which surprisal and attentional features derived from a Neural Machine Translation (NMT) model account for reading and production times of human translators. We find that surprisal and attention are complementary predictors of translation difficulty, and that surprisal derived from a NMT model is the single most successful predictor of production duration. Our analyses draw on data from hundreds of translators operating across 13 language pairs, and represent the most comprehensive investigation of human translation difficulty to date.
\end{abstract}

\section{Introduction}

\begin{figure}
         \includegraphics[width=\linewidth]{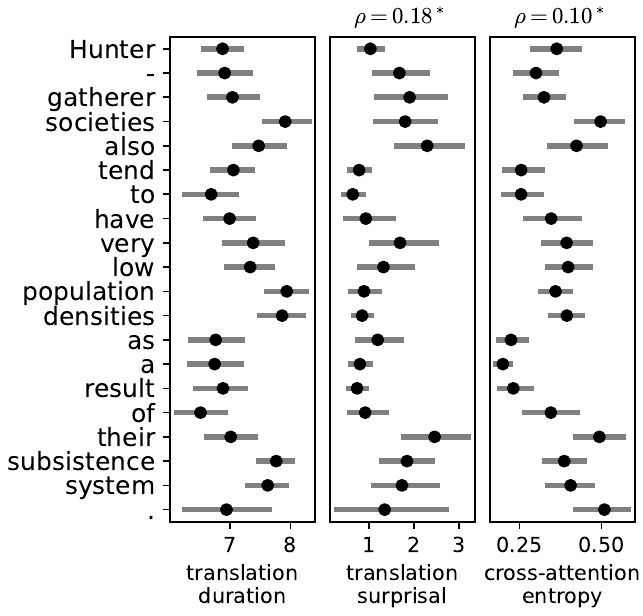}
         \caption{Translation surprisal and cross-attention entropy are predictive of average word translation duration, which is the time taken (in ms, log scale) to produce an aligned segment on the target side, divided by the number of source words aligned with this segment. Error bars show 95\% confidence intervals over 47 translations in five target languages: Chinese, Japanese, Hindi, German and Spanish. `*' marks significant Pearson correlations ($p <.05$) of predictors with translation duration (leftmost panel).}
         \label{fig:tgt-example}
\end{figure}

During the Nuremberg trials, a Soviet interpreter paused and lost her train of thought when faced with the challenge of translating the phrase ``Trojan Horse politics'', and the presiding judge had to stop the session~\cite{matasov20}. Translation difficulty rarely has such extreme consequences, but the process of translating any text requires a human translator to handle words and phrases that vary in difficulty. Translation difficulty can be operationalized in various ways~\cite{sun2015measuring}, and one approach considers texts to be difficult if they trigger translation errors~\cite{vanroy2019correlating}. Here, however, we focus on difficulty in cognitive processing, and consider a word or phrase to be difficult if it requires extended processing time. Figure~\ref{fig:tgt-example} illustrates how this notion of translation difficulty varies at the level of individual words.   Across this sample, words like ``societies'' and ``population'' are consistently linked with longer production times than words like ``result'' and ``tend''.

Processing times have been extensively studied by psycholinguists, but the majority of this work is carried out in a monolingual setting. Within the literature on translation, analysis of cognitive processing is most prominent within a small but growing area known as Translation Process Research (TPR). Researchers in this area aim to characterize the cognitive processes and strategies that support human translation, and do so by analyzing eye movements and keystrokes collected from translators~\cite{carl2012critt}. Here we build on this tradition and focus on three variables from the CRITT TPR-DB database~\cite{carl2016critt}: source reading time (\texttt{TrtS}), target reading time (\texttt{TrtT}), and translation duration (\texttt{Dur}).  Our analyses are relatively large in scale by the standards of previous work in this area, and we draw on data from 312 translators working across 13 language pairs. 

A central goal of our work is to bring translation process research into contact with modern work on Neural Machine Translation (NMT). Recent work in NLP has led to dramatic improvements in the multilingual abilities of NMT models~\cite{kudugunta2019investigating,aharoni2019massively}, and these models can support tests of existing psycholinguistic theories and inspire new theories.
Our work demonstrates the promise of NMT models for TPR research by testing whether surprisal and attentional features derived from an NMT model are predictive of human translation difficulty. Two of these predictors are shown in the right panels of Figure~\ref{fig:tgt-example}, and both are correlated with translation duration for the example sentence shown.

In what follows we introduce the surprisal and attentional features that we consider then evaluate the extent to which they yield improvements over baseline models of translation difficulty. We find that surprisal is a strong predictor of difficulty, which supports and extends previous psycholinguistic findings that surprisal predicts both monolingual processing~\cite{levy2008expectation,wilcox2023testing} and translation processing \cite{teich2020translation,wei2022entropy,carl2021information}. The attentional features we evaluate predict difficulty less well, but provide supplementary predictive power when combined with surprisal.

\section{Related work}

An extensive body of monolingual work has demonstrated that the 
pace of human reading is based on next-word predictability --- more contextually surprising words incur higher cognitive costs and are slower to process \cite{hale2001probabilistic,levy2008expectation}. The phenomenon is observed across a wide range of language usage, including reading \cite{monsalve2012lexical,smith2013effect,meister2021revisiting,shain_meister_pimentel_cotterell_levy_2022,wilcox2021targeted,wilcox2023testing}, listening and comprehension \cite{russo2020semantics,russo2022negative,kumar2022bayesian,yu2023neural}, speech \cite{jurafsky2003probabilistic,jaeger2006speakers,cohen2008phone,demberg2012syntactic,malisz2018dimensions,dammalapati2019expectation,pimentel2021surprisal}, typing \cite{chen2021factors} and code-switching \cite{calvillo2020surprisal}. Surprisal has also been proposed as a predictor of translation difficulty, but existing evaluations of this proposal are relatively limited \cite{teich2020translation,carl2021information,wei2022entropy,deilen4375757cognitive}.

Along with surprisal, contextual entropy has also been proposed as a predictor of translation difficulty \cite{teich2020translation,carl2021information,wei2022entropy}. Entropy is the expected surprisal of a translation distribution, and indicates the effort in resolving the uncertainty over all possible translation choices. However, the results of \citet{wei2022entropy} and \citet{carl2021information} suggest entropy is a weaker predictor of translation duration relative to surprisal. Entropy also shows limited evidence in predicting monolingual reading difficulty, despite being hypothesized to indicate a reader's anticipation of processing effort \cite{hale2003information,linzen2016uncertainty,lowder2018lexical,wilcox2023testing}. 

Previous studies have also explored whether attentional weights in language models can account for monolingual processing time~\cite{ryu2021accounting,oh2022entropy}.
Other NMT studies contend that the contextualisation of difficult tokens occurs in encoder self-attention \cite{tang2019encoders,yin2021context}. NMT models also exhibit higher cross-attention entropy with increasingly difficult tokens \cite{dabre2019recurrent,zhang2021modeling,lu2021attention}. Further, NMT's non-compositional processing of figurative expressions leads to paraphrasing, as opposed to literal translation \cite{dankers2022can}. 

Building on previous work, we focus on surprisal rather than entropy, and provide a comprehensive evaluation of the extent to which surprisal predicts translation difficulty. While we do not expect NMT difficulty to align perfectly with human difficulty, we propose several attentional features based on previous literature and test whether they contribute predictive power that goes beyond surprisal alone.\footnote{Code will be made available upon publication.}

\section{Surprisal: An information-theoretic account of translation processing}
\label{sec:surprisal}

In contrast with traditional monolingual surprisal, translation surprisal is based on a context that includes a complete sequence in a source language (SL) and a sequence of previously translated words in a target language (TL). Target words with high translation surprisal are hypothesized to require extended cognitive processing because they are relatively unpredictable in context. 

Relative to prior work, we aim to provide a more comprehensive and rigorous evaluation of the role of surprisal in translation processing. We consider predictors of translation difficulty based on both monolingual and translation surprisal, which we estimate using a large language model and a neural translation model respectively. 

\subsection{Monolingual surprisal}

The monolingual surprisal of a word or segment can be estimated from an autoregressive language model (LM). Let $\bs{w} = [w_1, ..., w_m]$ be a complete sequence of tokens (e.g., a full sentence), and let $\bs{w}_{\bs{i}}$ denote a segment of $\bs{w}$ where $\bs{i} \subseteq \{1, ..., m\}$ denotes the token indices.
The surprisal of $\bs{w}_{\bs{i}}$ is the sum of negative log-probabilities of each token $w_i$ conditioned on the preceding context $\bs{w}_{<i}$:

\begin{equation}
    \label{eq:lm-surp}
    s_{\mathrm{lm}}(\bs{w}_{\bs{i}}) = \sum\limits_{i \in \bs{i}}-\log p_{\mathrm{lm}}(w_{i}|\bs{w}_{<i}),
\end{equation}

\noindent
where $p_{\mathrm{lm}}$ denotes the LM distribution. Note that this definition of surprisal applies to both words and segments, which consist of a sequence of tokens defined under a model tokenization scheme.\footnote{$\bs{w}_{\bs{i}}$ is also defined in a way that allows difficulty prediction of non-contiguous segments.}  

Given the well-established link between $s_{\mathrm{lm}}$ and reading time \cite{monsalve2012lexical,smith2013effect,shain_meister_pimentel_cotterell_levy_2022,wilcox2021targeted}, we test whether $s_{\mathrm{lm}}$ predicts reading times for source and target texts when participants are engaged in the act of translation.
Previous work also establishes links between surprisal and language production in tasks involving speech
\cite{dammalapati2021effects}, oral reading \cite{klebanov2023dynamic}, code-switching  \cite{calvillo2020surprisal} and typing \cite{chen2021factors}. Building on this literature, we also test whether monolingual surprisal predicts the time participants take to type their translations. 

\subsection{Translation surprisal}
The surprisal of a translation can be obtained from the distribution of a neural machine translation model, $p_{\mathrm{mt}}$, conditioned on a source sequence and previously translated tokens. Let $\bs{x}$ and $\bs{y}$ be a pair of parallel sequences, and let $\bs{y}_{\bs{j}}$ be a segment of $\bs{y}$ with token indices $\bs{j}$ and $\bs{y}_{<j}$ be the preceding context of each token $y_{j}$. The translation surprisal of segment $\bs{y}_{\bs{j}}$ is defined as
\begin{equation}
    \label{eq:mt-surp}
    s_{\mathrm{mt}}(\bs{y}_{\bs{j}}) =  \sum\limits_{j \in \bs{j}}-\log p_{\mathrm{mt}}(y_{j}|\bs{x}, \bs{y}_{<j}).
\end{equation}
\noindent
We will compare translation surprisal and monolingual surprisal as predictors of target-side reading time and production duration. We expect translation surprisal to be the more successful predictor because this measure incorporates context on both the source and target sides, and because NMT models are trained specifically for translation. Translation surprisal, however, cannot be used to predict source-side reading time because the model must encode the entire source sequence before providing an output distribution.

\section{Predicting translation difficulty using NMT attention}
\label{sec:attention}

State-of-the-art NMT models rely on the transformer architecture and incorporate three kinds of attention: encoder self-attention, cross-attention and decoder self-attention.\footnote{Cross-attention is sometimes known as encoder-decoder attention.} We consider all three sets of attention weights as potential predictors of translation difficulty. By some accounts, reading, transferring and writing are three distinct stages in the human translation process \cite{shreve1993there,macizo2004translation,macizo2006reading,shreve2017aspects}, and we propose that the three attentional modules roughly align with these three stages.

Let $\bs{x} = [x_1, ..., x_m]$ and $\bs{y} = [y_1, ..., y_n]$ denote parallel source and target sequences,  with $\bs{m} = \{1,...,m\}$ and $\bs{n} = \{1,...,n\}$ as their token indices. 
Let $\bs{u}$ and $\bs{v}$ denote segments of $\bs{x}$ and $\bs{y}$ respectively such that $\bs{u} = \bs{x}_{\bs{i}}$ and $\bs{v} = \bs{y}_{\bs{j}}$, where the indices $\bs{i} \subseteq \bs{m}$ and $\bs{j} \subseteq \bs{n}$. Note that $\bs{x}$ and $\bs{y}$ do not include special tokens (e.g.,\ the end-of-sequence tag  \texttt{eos}) added to the sequences under the  NMT's tokenization scheme. We additionally define 
$\bs{\nt{u}} = \bs{x}_{\bs{\nt{i}}}$ and $\bs{\nt{v}} = \bs{y}_{\bs{\nt{j}}}$ as the contexts where the corresponding segments are excluded, i.e., $\bs{\nt{i}} = \bs{m} \setminus \bs{i}$ and $\bs{\nt{j}} = \{1, ..., \max(\bs{j})\} \setminus \bs{j}$.
In contrast to $\bs{\nt{i}}$, $\bs{\nt{j}}$ only includes token indices from the context preceding $\bs{v}$. We define $\bs{\nt{j}}$ in this way because the decoder typically does not have access to future tokens when generating translations.

We consider two kinds of attentional features. The first captures the total attentional flow from segment $\bs{w}$ to segment $\bs{z}$:
\begin{equation}
\label{eq:att}
 \mathrm{flow}(A, \bs{w}, \bs{z}) = \sum\limits_{l \in \bs{l}} \sum\limits_{k \in \bs{k}}  a_{kl}, \\
\end{equation}
where $A$ is an attention matrix, $\bs{k}$ and $\bs{l}$ are token indices of $\bs{w}$ and $\bs{z}$, and
$a_{kl}$ is the attention from the token at index $k$ to the token at index $l$.

The second kind of attentional feature sums the entropies of all attentional distributions from 
segment $\bs{w}$ to segment $\bs{z}$:
\begin{gather}
\label{eq:ent}
    H(A, \bs{w}, \bs{z}) = \sum\limits_{k \in \bs{k}} \sum\limits_{l \in \bs{l}} - \hat{a}_{kl} \log \hat{a}_{kl} \\
    \hat{a}_{kl} = \frac{a_{kl}}{\lVert \bs{a}_{k\veryshortarrow\bs{l}} \rVert}, \nonumber
\end{gather}
\noindent
where again $\bs{k}$ and $\bs{l}$ are the indices of $\bs{w}$ and $\bs{z}$. We renormalize $a_{kl}$ along $\bs{a}_{k\veryshortarrow\bs{l}}$, the attention vector at $k$ with target indices $\bs{l}$.\footnote{$\ell_{1}$ normalization} This step allows us to filter out attention to special tokens (e.g.\  \texttt{eos}) when considering the entropy of a distribution.

Using Equations~\ref{eq:att} and \ref{eq:ent}, we define in the following 6 attentional features as candidate predictors of source-side reading time and 5 features for target-side reading time and production duration. 

\subsection{Predicting source text difficulty}
\label{sec:src-diff}
\textbf{Encoder attention}. We propose four features extracted from encoder attention that are inspired by \citet{dankers2022can}, who find that when producing a paraphrase rather than a literal translation  of a figurative expression, an NMT encoder tends to
allocate more attention from the phrase to itself, while reducing attention directed to and received from the context. This result is relevant 
to our goal of predicting translation difficulty 
because  paraphrasing is known to be more effortful than literal translation \cite{balling2014evidence,schaeffer2014measuring,rojo2015translation,carl2017translation}. 
 
 Given an encoder self-attention matrix $A^e$, we use Equation~\ref{eq:att} to define features that capture the total attentional flow from $\bs{u}$ to $\bs{u}$, to its context $\bs{\nt{u}}$, and to the \texttt{eos} tag:
\begin{align}
\nonumber
    f^{\mathrm{e}}_{u,u} &= \mathrm{flow}(A^{\mathrm{e}}, \bs{u}, \bs{u})  \\  \nonumber
    f^{\mathrm{e}}_{u,\nt{u}}  &= \mathrm{flow}(A^{\mathrm{e}}, \bs{u}, \bs{\nt{u}}) \\ \nonumber
    f^{\mathrm{e}}_{u,\texttt{eos}} &= \mathrm{flow}(A^{\mathrm{e}}, \bs{u}, \texttt{eos}). 
\end{align}

\noindent
In line with \citet{dankers2022can}, we hypothesize that harder-to-translate segments direct more attention to themselves and less to their contexts. To examine if the model also reduces attention flow from context to $\bs{u}$, we further define \att{e}{\nt{u}}{u}:
\begin{equation}
    f^{\mathrm{e}}_{\nt{u},u} = \mathrm{flow}(A^{\mathrm{e}}, \bs{\nt{u}}, \bs{u}). \nonumber
\end{equation}

The NMT encoder relies on attention to relevant context to disambiguate input meanings and to resolve anaphoric pronouns \cite{tang2019encoders,yin2021context}. These words may take longer to read, as more time is required to determine pronoun referents and word senses. Ambiguous words also contribute to low attentional entropy  \cite{tang2019encoders}. To characterize this feature of $\bs{u}$, we compute \ent{e}{u}{x}, the overall attentional entropy based on Equation~\ref{eq:ent}:
\begin{equation}\nonumber
    H^{\mathrm{e}}_{u,x} = H(A^{\mathrm{e}}, \bs{u}, \bs{x}). \\ \nonumber
\end{equation}
\noindent
We hypothesize that low \ent{e}{u}{x} predicts longer reading time of $\bs{u}$.

\medskip\noindent
\textbf{Cross-attention}. Cross-attention allows information to pass from encoder to decoder and establishes rough alignments between input and output tokens in an NMT model \cite{alkhouli2018alignment,li2019word}. However, it is unclear from previous work whether more attention weight received from the target sequence contributes to harder or easier translation.  \citet{tu2016modeling} and \citet{mi2016coverage} show that increased attention received by part of the source text is related to \textit{over-translation}, a phenomenon in which the model focuses too much on some parts of the input and neglects others when generating a translation. In contrast, \citet{dankers2022can} demonstrate that paraphrasing a figurative expression instead of literal translation reduces attention to corresponding source tokens.

On one hand, source tokens that receive more attention and stronger alignments are deemed more important by the model; on the other, the same phenomenon corresponds to literal translation, which is easier than paraphrasing. Nevertheless, both studies show that cross-attention flow to $\bs{u}$ is pertinent to our goal of characterizing translation difficulty.
Given a cross-attention matrix $A^{\mathrm{c}}$, we define the attentional flow from $\bs{y}$ to $\bs{u}$ as:
\begin{equation}
    f^{\mathrm{c}}_{y,u} = \mathrm{flow}(A^{\mathrm{c}}, \bs{y}, \bs{u}), \nonumber
\end{equation}
and test \att{c}{y}{u} as a predictor of source reading time.

\subsection{Predicting target text difficulty}
\label{sec:tgt-diff}
\textbf{Cross-attention}.
When paraphrasing non-literal nouns, NMT also tends to shift cross-attention weight away from these nouns, focusing instead on surrounding tokens and source \texttt{eos} \cite{dankers2022can}. 
We therefore hypothesize that the translation difficulty associated with target segment $\bs{v}$ may be predicted by the total attention directed from $\bs{v}$ to source \texttt{eos}:

\begin{equation}
    f^{\mathrm{c}}_{v,\texttt{eos}} = \mathrm{flow}(A^{\mathrm{c}}, \bs{v}, \texttt{eos}). \nonumber
\end{equation}

Prior work attributes cross-attention uncertainty to lack of confidence in translation outputs and less informative source inputs~\cite{dabre2019recurrent,zhang2021modeling,lu2021attention}.
Following these studies, we hypothesize that higher uncertainty in cross-attention indicates less confident alignment with source tokens and therefore predicts increased translation difficulty. We measure alignment uncertainty between  $\bs{v}$ and the source sequence $\bs{x}$ as:
\begin{equation}
  H^{\mathrm{c}}_{v,x} = H(A^{\mathrm{c}}, \bs{v}, \bs{x}). \nonumber
\end{equation} 

\medskip\noindent
\textbf{Decoder attention}. While previous work on NMT models has considered encoder and cross-attention in depth, decoder self-attention has received less investigation. \citet{yang2020sub} demonstrates that the role of decoder self-attention is to ensure translation fluency.
Relative to encoder attention and cross attention, decoder attention aligns less well with human annotations, and contributes less to improving NMT performance when regularized with human annotations \cite{yin2021context}.

For completeness, however, we consider three decoder attentional features that parallel those introduced in Section~\ref{sec:src-diff} for encoder self-attention. Despite their similarity, these attentional features are evaluated against different behavioral measures. The encoder features are treated as candidate predictors of source reading time, and the decoder features are used to predict target reading and production time. If $A^{\mathrm{d}}$ is the decoder attention matrix, the following features capture attention flow from $\bs{v}$ to itself and to the preceding context, as well as self-attention entropy:
\begin{align}
\nonumber
    f^{\mathrm{d}}_{v,v} &= \mathrm{flow}(A^{\mathrm{d}}, \bs{v}, \bs{v})  \\  \nonumber
    f^{\mathrm{d}}_{v,\nt{v}} &= \mathrm{flow}(A^{\mathrm{d}}, \bs{v}, \bs{\nt{v}}) \\\nonumber
    H^{\mathrm{d}}_{v,\overleftarrow{v}} &= H(A^{\mathrm{d}}, \bs{v}, \bs{\overleftarrow{v}}).
\end{align}

\noindent
As mentioned earlier, a decoder does not attend to future tokens, including target-side \texttt{eos}. Decoder features analogous to \att{e}{u}{\texttt{eos}} and \att{e}{\nt{u}}{u} are  not possible for this reason. Attentional entropy \ent{d}{v}{\overleftarrow{v}} is 
computed over $\bs{\overleftarrow{v}}$,  which includes all target-side tokens up to the rightmost token in $\bs{v}$.

\begin{table*}
\footnotesize
\setlength{\tabcolsep}{4.5pt}
\centering
\begin{tabular}{cccccccc}
\toprule
 & Study & \multicolumn{3}{c}{Token} & \multicolumn{3}{c}{Segment}  \\
&  &  \texttt{TrtS} & \texttt{TrtT} &  \texttt{Dur} & \texttt{TrtS} & \texttt{TrtT} &  \texttt{Dur}  \\
\midrule \noalign{\vskip 0.7mm} 
 en $\rightarrow$ da & \makecell{ACS08 \cite{sjorup2013cognitive}, \\BD13, BD08 \cite{dragsted2010coordination}}  & 5305 & 5320 & 6176 & 4121.0 & 4203.0 & 4779.0
 \\ \noalign{\vskip 0.7mm} 
en $\rightarrow$ de &SG12 \cite{nitzke2019problem}  & 3691 & 3956 & 	4534 &  2991.0 & 3243.0 & 3589.0

\\\noalign{\vskip 0.7mm} 
en $\rightarrow$ es & BML12 \cite{mesa2014gaze}  & 4067	 & 4020	 & 8280	 & 3555.0 & 3330.0 & 6072.0\\\noalign{\vskip 0.7mm} 
en $\rightarrow$  hi &  NJ12 \cite{carl2016critt}  & 4717 & 	4828 & 	5205 & 2917.0 & 2851.0 & 2933.0 \\\noalign{\vskip 0.7mm} 
en $\rightarrow$  ja & ENJA15 \cite{carl2016japanese}  & 6806 & 8329 & 	2168 & 4263 & 4299 & 2130 \\\noalign{\vskip 0.7mm} 
en $\rightarrow$ nl &  ENDU20 \cite{vanroy2021syntactic}  & 0 & 0 & 7814 & 0 & 0 & 6318 \\\noalign{\vskip 0.7mm} 
en $\rightarrow$  pt & JLG10 \cite{alves2013investigating}  & 0 & 0 & 2443 & 0 & 0 & 2217
 \\\noalign{\vskip 0.7mm} 
en $\rightarrow$ zh & \makecell{ RUC17, STC17 \cite{carl2019machine},\\ CREATIVE \cite{vieira2023translating}}  & 8949 & 7934 &	3876 & 6097 & 5922 & 3925
 \\\noalign{\vskip 0.7mm} 
da $\rightarrow$ en & LWB09 \cite{jensen2009effects}  & 3844 & 4177 & 5327 &  3445 & 3493 & 4315 \\\noalign{\vskip 0.7mm} 
fr $\rightarrow$ pl & DG01 \cite{plonska2016problems}  & 0 & 0 & 17041 & 0 & 0 & 13283
\\\noalign{\vskip 0.7mm} 
pt $\rightarrow$ en & JLG10  & 0 & 0 & 2053	 & 0 & 0 & 1876 \\\noalign{\vskip 0.7mm} 
pt $\rightarrow$ zh & MS13 \cite{schmaltz2016cohesive} & 1011 & 830 & 203	& 781 & 755 & 237 \\\noalign{\vskip 0.7mm} 
zh $\rightarrow$ pt & MS13  & 1210 & 1237 & 1509 & 1101 & 1027 & 1209 \\\noalign{\vskip 0.7mm} 
\bottomrule
\end{tabular}
\caption{ Data drawn from studies in CRITT-TPRDB with the number of valid samples ($\geq$ 20ms) per segmentation level. }
\label{tab:data-size}
\end{table*}

\section{Methods}
\label{sec:modeling}
\subsection{Experimental setup}
\textbf{Data.} Our measures of translation difficulty are derived from the CRITT Translation Process Research Database (TPR-DB) \cite{carl2016critt}.  We focus on three behavioral measures  of translation difficulty: source text reading time (\texttt{TrtS}), target text reading time (\texttt{TrtT}) and translation production duration (\texttt{Dur}). Both reading time measures are based on eye-tracking data, and the translation production measure is based on keylogging data. The data set is organized in terms of words, segments, and sentences, and we carry out separate analyses at the word and segment levels.

 In CRITT TPR-DB, word and segment boundaries and alignments are provided by human annotators. For consistency, we remove alignments of words and segments that cross sentence boundaries.  Following  \citet{carl2021translation}, we filter values of \texttt{TrtS}, \texttt{TrtT} and \texttt{Dur} lower than 20ms. The remaining values are log scaled. We analyze data from 17 public studies available from the public database.\footnote{The translation studies selected from the \href{https://sites.google.com/site/centretranslationinnovation/tpr-db/public-studies}{TPR database} exclude data sets where many alignments cross sentence boundaries, or that contain too many errors (e.g., missing values and inconsistent sentence segmentations) across tables. } These studies represent 13 different language pairs, and each study includes data from an average of 18 human translators. The studies included along with the size of each  one are summarized in Table~\ref{tab:data-size}. For cross-validation, we divide the samples into 10 folds. To ensure that all predictions are evaluated using previously unseen sentences, we randomly sample test data at the sentence level, which means that the source sentences in train and test partitions do not overlap.  

\medskip\noindent
\textbf{LM and NMT models.} We use mGPT \cite{shliazhko2022mgpt}, a multilingual language model, to estimate monolingual surprisal ($s_{\mathrm{lm}}$).\footnote{\href{https://huggingface.co/ai-forever/mGPT}{mGPT checkpoint}} To compute translation surprisal ($s_{\mathrm{mt}}$), we use NLLB-200's 600M variant, which is distilled from a much larger 54.5B Mixture-of-Experts model \cite{costa2022no}.\footnote{\href{https://huggingface.co/facebook/nllb-200-distilled-600M}{NLLB-200 checkpoint}} 
We compute attentional features for each of 16 heads across 12 layers, then average across heads and layers to create the final set of attentional features for our analyses.

\medskip\noindent
\textbf{Normalization.}
Sections~\ref{sec:surprisal} and \ref{sec:attention} describe feature definitions that are sums over a sequence of tokens, which makes it crucial to control for segment length when predicting reading time and production duration. All surprisal values are therefore normalized by the lengths of the input segments, $\bs{w}_{\bs{i}}$ and $\bs{y}_{\bs{j}}$. To normalize attentional features, we first calculate dummy feature values by replacing $a_{lk}$ in Equation~\ref{eq:att} and \ref{eq:ent} with uniform attention values (i.e., $a_{lk}=1/|k|$ where $|k|$ is the length of the attention vector). A normalized attentional feature is defined as the ratio of the raw feature value (defined in Section~\ref{sec:attention}) to its dummy value. 

\medskip\noindent
\textbf{Control features.}
Although we are most interested in surprisal and attentional features as predictors of translation difficulty, other simple features may also predict difficulty. In particular, longer segments, low-frequency segments, and segments towards the beginning of a sentence might be systematically more difficult than shorter segments, high-frequency segments, and segments towards the end of a sentence. We therefore include segment length, average unigram frequency\footnote{\url{https://pypi.org/project/wordfreq/}} (log scaled) and average position quantile as control features in all models, where both averages are computed over all tokens belonging to a segment.

\medskip\noindent
\textbf{Linear models.} Following previous studies, we use linear models to evaluate the predictive power of both surprisal and attentional features. To predict translation difficulty for all languages, we use a mixed model that includes language pair and participant id as random effects. In addition, we fit individual linear regression models for the four language pairs (en $\rightarrow$ da, en $\rightarrow$ de, en $\rightarrow$ hi, and en $\rightarrow$ zh) for which we have most data. 

\subsection{Log-likelihood delta as an estimation of predictive power}
Following previous psycholinguistic studies \cite{goodkind2018predictive,kuribayashi2021lower,wilcox2020predictive,wilcox2023testing,de2023scaling} we evaluate our models using log-likelihood of each sample in the held-out data. To assess the predictive power of a feature, we train a model with the feature of interest in addition to all control features, and compare against a baseline model which includes only the control features. The contribution of the predictor feature is then measured as the difference in log-likelihood of the held-out test data ($\Delta \mathrm{ llh}$) between the two models. A positive $\Delta \mathrm{ llh}$ indicates added predictive power from the feature relative to the baseline model, whereas $\Delta \mathrm{ llh} \leq 0$ means that we have no evidence for the effect of the feature on reading and production times.\footnote{ \citet{wilcox2023testing} point out that $\Delta \mathrm{ llh}$ may be $\leq 0$ because of overfitting, or because the relationship between the predictor feature and the target variable is not adequately captured by the model class used (in our case, linear models).} Like \citet{wilcox2023testing}, we test if $\Delta \mathrm{ llh} > 0$ is significant across held-out samples with a paired permutation test.\footnote{Our test uses 1000 random permutations.}

\section{Results}
\label{sec:result}
\subsection{Surprisal and attentional features}
\label{sec:result_inde}

\begin{figure}
     \centering
     \includegraphics[width=0.47\textwidth]{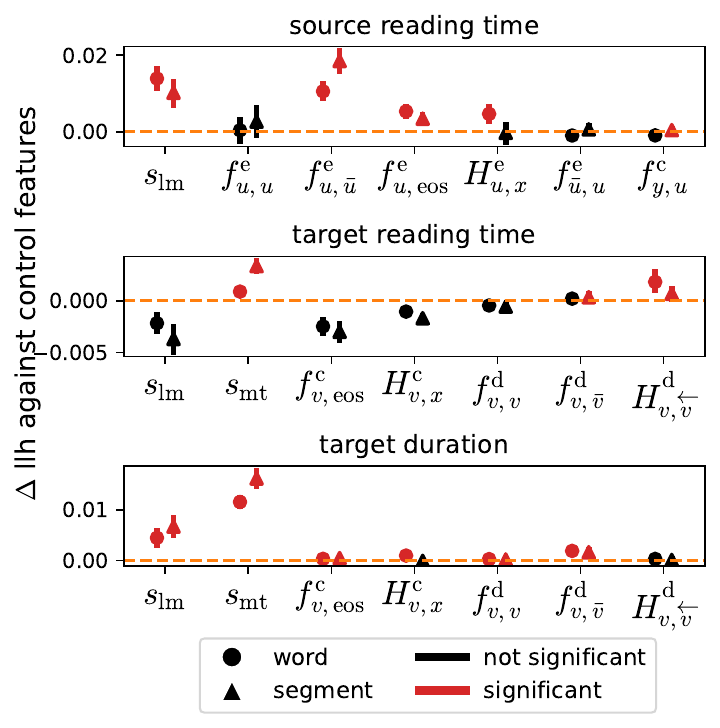}
     \caption{$\Delta \mathrm{ llh}$ plotted against control features. 95\% confidence intervals are estimated based on the 10 cross-validation folds.}
     \label{fig:llh-delta}
\end{figure}

Figure~\ref{fig:llh-delta} 
shows $\Delta \mathrm{ llh}$ of surprisal predictors and attentional features at both word and segment levels.
Data points shown in red indicate predictors that are statistically significant (p < .05) relative to the baseline model. When fitted on all language pairs, $s_{\mathrm{lm}}$ is a significant predictor of source text reading time and target production duration, but not target reading time. On the target side, $s_{\mathrm{mt}}$  is the best overall predictor of difficulty. Target reading time has fewer significant predictors than does target production duration, and therefore appears to be harder to predict. From here on, we restrict our analyses to features that are significant at at least one of the two segmentation levels.

\begin{figure*}
     \centering
     \includegraphics[width=0.9\textwidth]{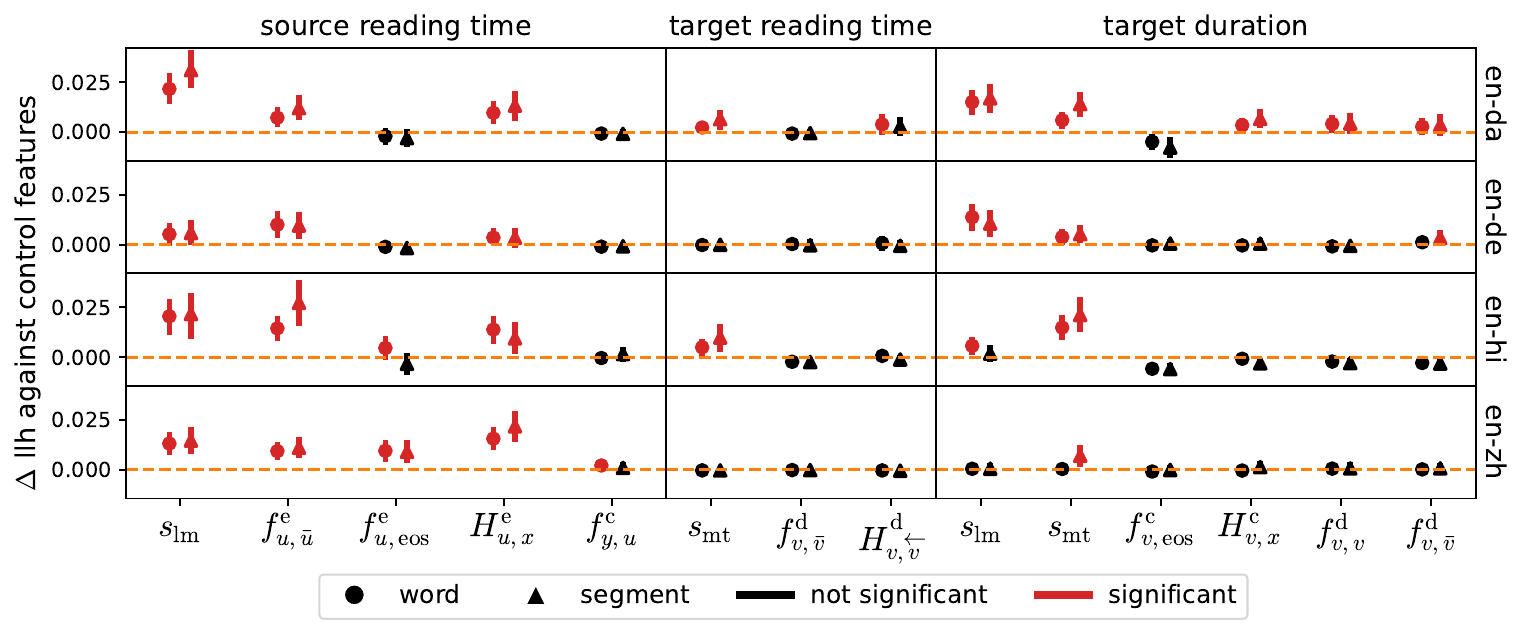}
     \caption{$\Delta \mathrm{ llh}$ by language pairs with features that are significant for at least one of the word and segment levels in Figure~\ref{fig:llh-delta}.}
     \label{fig:llh-delta-lpair}
\end{figure*}

Figure~\ref{fig:llh-delta-lpair} shows analogous results for four individual language pairs.
Surprisal features $s_{\mathrm{lm}}$ and $s_{\mathrm{mt}}$ remain strong predictors for reading time and target production duration in general. Among the attentional features, \att{e}{u}{\nt{u}} and \ent{e}{u}{x} most consistently predict source reading time across language pairs and segmentation levels. However, the predictions of target reading time and duration by attentional features are less consistent --- despite predicting en $\rightarrow$ da target difficulty, most attentional features fail to contribute in other language pairs. One possible reason is that these features overfit to small samples of individual language pairs, compared to a mixed model that is fitted on a much larger data set including all language pairs.

\subsection{Attention is supplementary to $s_{\mathrm{lm}}$ and $s_{\mathrm{mt}}$}

Our results so far confirm that both $s_{\mathrm{lm}}$ and $s_{\mathrm{mt}}$ individually predict translation difficulty, whereas attentional features on their own are less consistent. We next ask if the attentional features that proved significant in Section~\ref{sec:result_inde} provide supplementary predictive power when combined with $s_{\mathrm{lm}}$ and $s_{\mathrm{mt}}$. To predict source reading time, we train models that include control features, $s_{\mathrm{lm}}$ and one attentional feature. For target difficulty, the models are trained with control features, $s_{\mathrm{mt}}$ and an attentional feature. We then calculate two variants of $\Delta \mathrm{ llh}$ for these models; the first compares against the baseline model, and the second compares against a model that is trained on control features and either $s_{\mathrm{lm}}$ or $s_{\mathrm{mt}}$. 

\begin{figure*}
     \centering
     \includegraphics[width=1\textwidth]{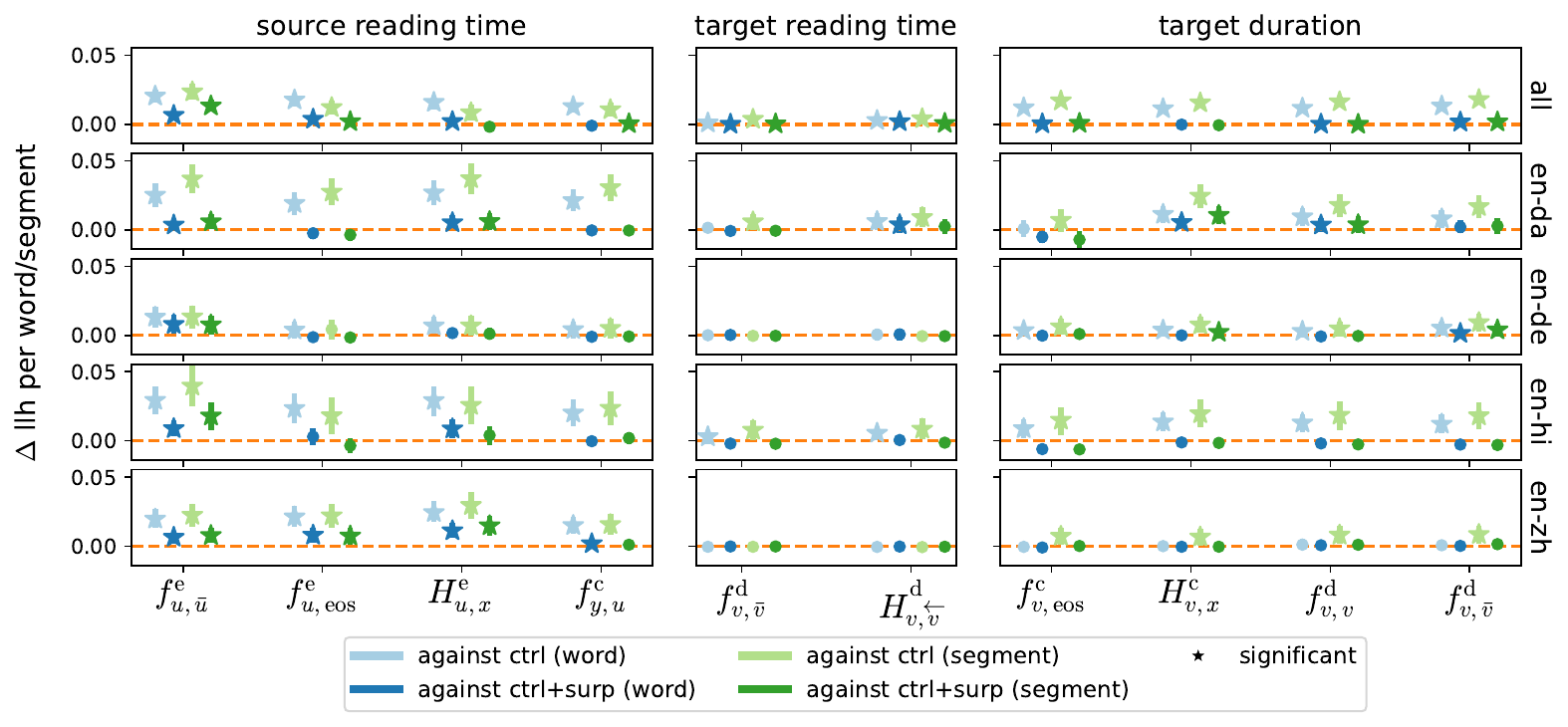}
     \caption{$\Delta \mathrm{ llh}$ with one feature in addition to $s_\mathrm{lm}$/$s_\mathrm{mt}$ and control features in training. Light and dark colored stars indicate $\Delta \mathrm{ llh}$ to be significantly above zero when compared against control features and control features with surprisal respectively. 
     For example, \att{c}{v}{\texttt{eos}} significantly contributes in addition to $s_\mathrm{mt}$ in predicting target duration at both word and segment levels,
     but \ent{c}{v}{x} does not contribute predictive power beyond $s_\mathrm{mt}$. 
     }
     \label{fig:llh-delta-att}
\end{figure*}

We repeat the same significance tests as before, and the results are shown in Figure~\ref{fig:llh-delta-att}. For the entire data set (top row of Figure~\ref{fig:llh-delta-att}), models with the addition of individual attentional features predict translation difficulty better than those trained with surprisal and control features only (except \ent{c}{v}{x}). Again, however, these results are weaker for individual language pairs. 

\begin{figure*}
     \centering
     \begin{subfigure}[b]{0.33\textwidth}
     \includegraphics[width=\linewidth]{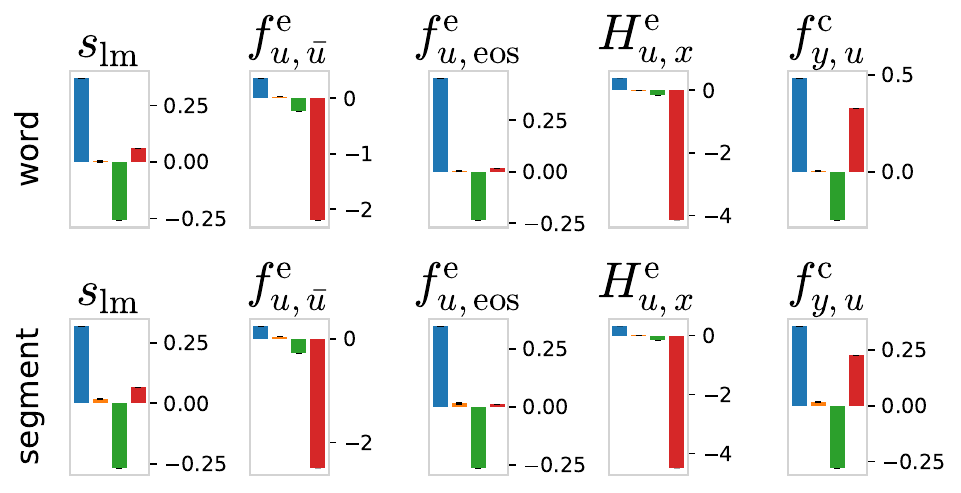}
     \caption{Source reading time}
     \label{fig:trts-coef}
     \end{subfigure}%
     \hspace{-1.9cm}
    \begin{subfigure}[b]{0.48\textwidth}
        \includegraphics[width=\linewidth]{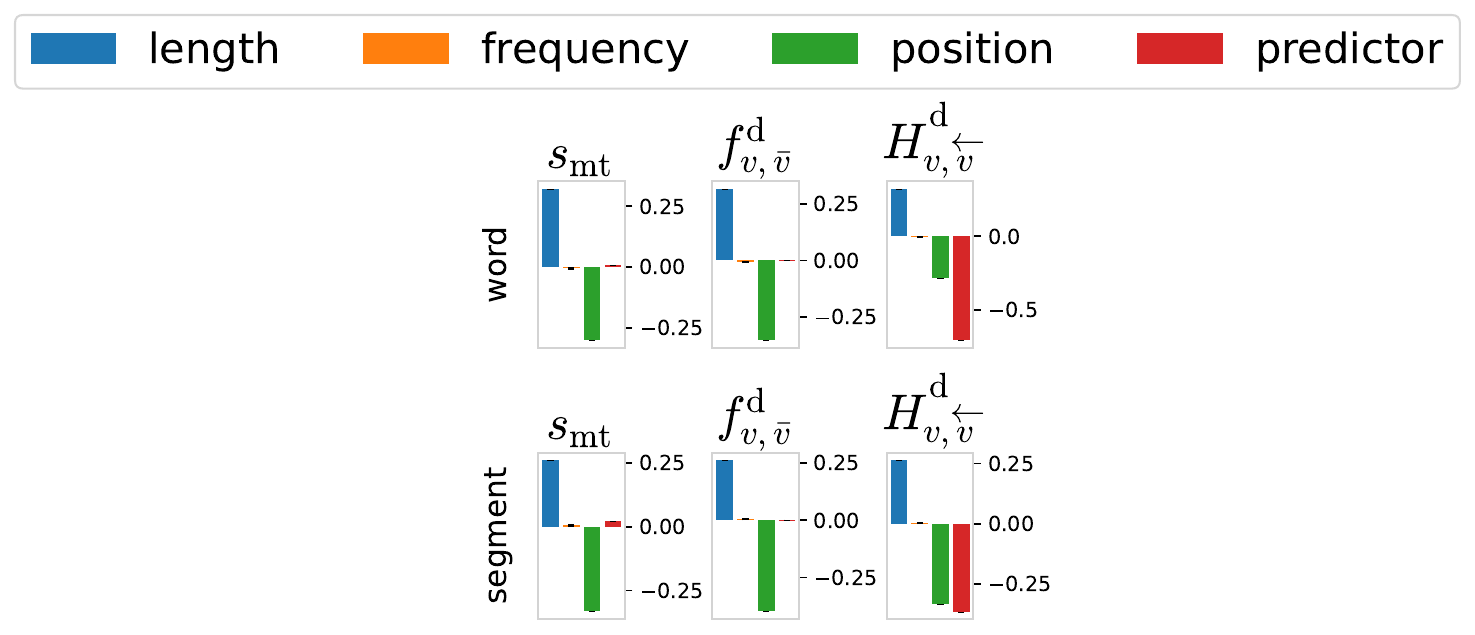}
         \caption{Target reading time}
     \label{fig:trtt-coef}
    \end{subfigure}%
    \hspace{-1.8cm} 
    \begin{subfigure}[b]{0.41\textwidth}
        \includegraphics[width=\linewidth]{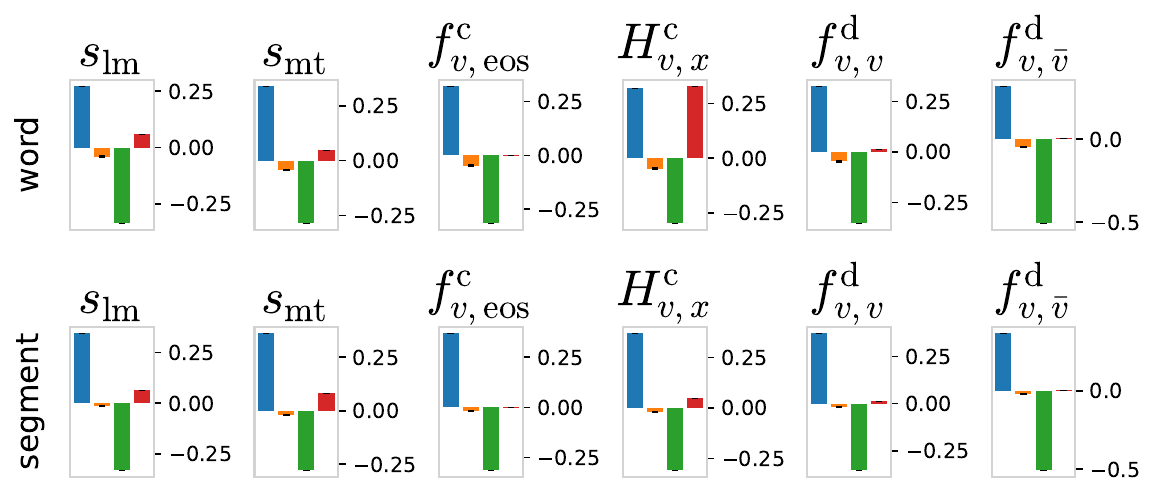}
         \caption{Target production duration}
     \label{fig:dur-coef}
    \end{subfigure}
     \caption{Predictor coefficients for linear models that include all three control features (length, frequency, position) along with one additional predictor (either surprisal or an attentional feature). Confidence intervals are plotted but are hard to see because they are so narrow.}
     \label{fig:coef}
\end{figure*} 

\subsection{Predictor coefficients}

Thus far we have only demonstrated the predictive power of surprisal and attentional features. To enable conclusions about the nature of the relationship between individual features and translation difficulty, Figure~\ref{fig:coef} shows average mixed model coefficients over data folds. The coefficients plotted support conclusions about the direction and effect size of the relationship between each predictor and translation difficulty, but the bar heights may not reflect predictive power, which has been indicated previously by $\Delta \mathrm{ llh}$.\footnote{We tested for collinearity by calculating variance inflation factors (VIF) for each set of features. The highest VIF is 2.2, which is below the thresholds (2.5/3) recommended by \citet{szmrecsanyi2006morphosyntactic} and \citet{zuur2010protocol}.} In general, segments are more difficult  when they are longer and occur earlier in the sequence. Rare words in general take longer to read and produce, as our baseline models consistently converge to negative coefficients for frequency (not shown in the figure). However, with the addition of surprisal and attentional features, the frequency effect for rare source words is reversed, whereas rare targets still require more attention and take longer to produce. As expected, increases in $s_{\mathrm{lm}}$ and $s_{\mathrm{mt}}$ are associated with increased reading time and production duration.

On the source side, the coefficents related to encoder self-attention indicate that harder-to-translate source texts direct less attention to context and more to \texttt{eos}, which reduces their entropy. 
Difficult source words are also singled out as important by having more incoming cross-attention from the target sequence.

On the target side, harder translations tend to show slight increases in cross-attention to source \texttt{eos}, and show more diffuse attention across the source sequence.\footnote{Mean coefficients of \att{c}{v}{\texttt{eos}} for token and segment are .001 and .002 respectively.} 
Our results thus support \citet{dankers2022can}'s claim that paraphrases show increased attention to \texttt{eos} and take longer to produce than literal translations.

Figures~\ref{fig:trtt-coef} and \ref{fig:dur-coef} also suggest that harder translations have more informative decoder attention, and direct more attention to themselves and the context. These results imply reduced attention to \texttt{bos}, the initial token of a translation sequence that conveys the target language to the NLLB model. 

\section{Discussion}

Section~\ref{sec:result_inde} showed that monolingual surprisal predicts source reading time, but that translation surprisal is a more consistent predictor of target reading time and production duration. On its own, NMT attention also predicts translation difficulty to some degree, but the most accurate predictions are achieved by combining surprisal and attentional features.

\subsection{Psycholinguistic implications}

Our results support previous findings that surprisal predicts translation difficulty \cite{wei2022entropy,teich2020translation,carl2021information}.
Surprisal has several justifications as a cognitive difficulty metric~\cite{levy2013memory,futrell2022information}, and one approach interprets surprisal as a measure of a shift in cognitive resource allocation. On this account, higher translation surprisal indicates that more effort is needed to shift cognitive resources to the word ultimately selected~\cite{wei2022entropy}.

\citet{teich2020translation}
suggests that translators aim for translations that are both faithful to the source ($p_\mathrm{mt}(t|s)$ is high) and fluent ($p_\mathrm{lm}(t)$ is high).  
These goals do not always align, and capture two different translation strategies --- literal translation optimizes MT probability at the expense of LM probability, whereas figurative translation prioritizes the latter. Although increases in $s_\textrm{lm}$ and $s_\textrm{mt}$ both predict increased target difficulty, our data reveals that these predictors have a weak but significant negative correlation ($p <.001$) at both token ($\rho=-.053$) and segment ($\rho=-.079$) levels. We therefore find quantitative support for a trade-off between fidelity and fluency~\cite{muller-etal-2020-domain}.

\subsection{Translatability by parts of speech}
\begin{figure*}
     \centering
     \begin{subfigure}[b]{0.67\textwidth}
     \includegraphics[width=\linewidth]{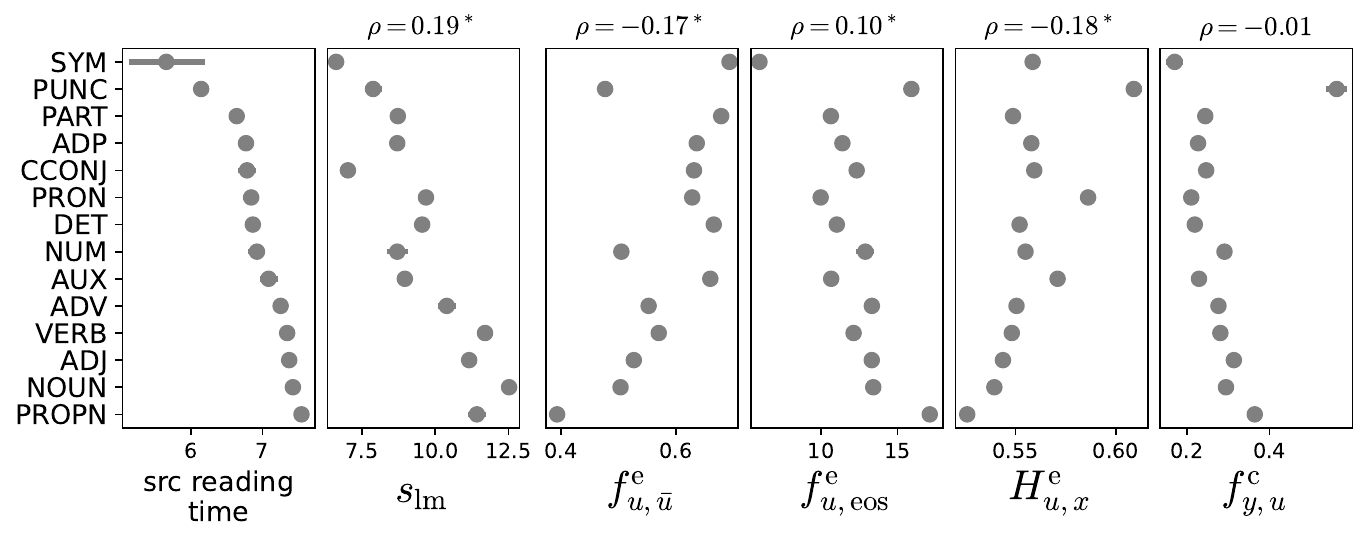}
     \caption{Source word reading time and predictors}
     \label{fig:src-POS}
     \end{subfigure}
    \begin{subfigure}[b]{0.9\textwidth}
        \includegraphics[width=\linewidth]{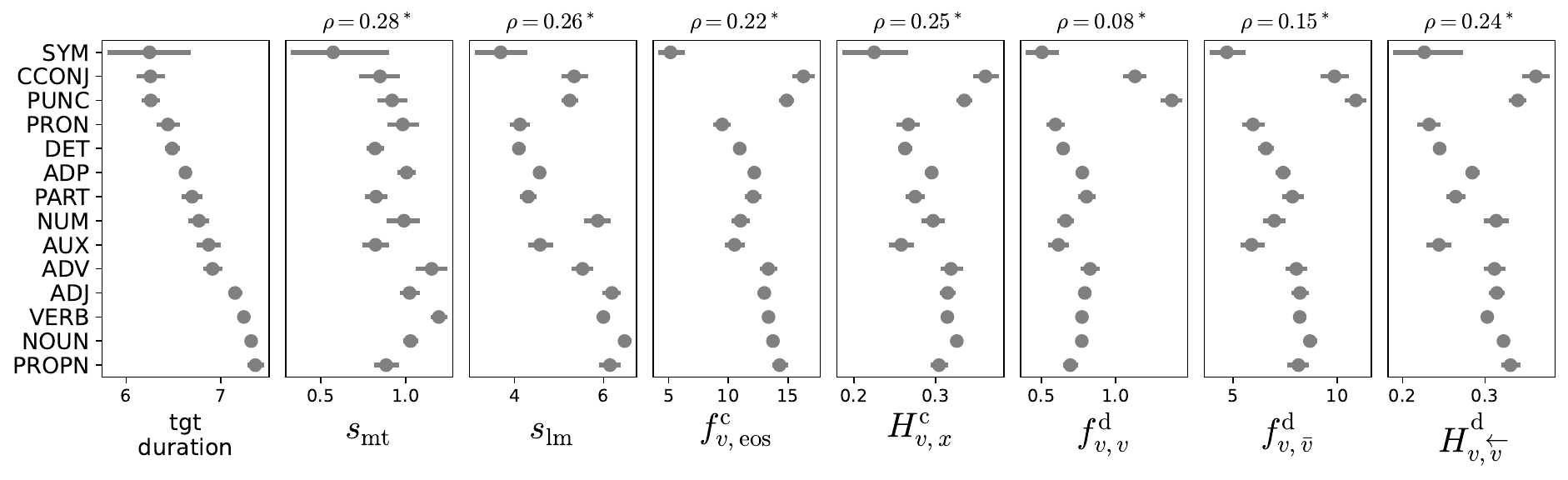}
         \caption{Average translation duration and predictors}
     \label{fig:tgt-POS}
    \end{subfigure}
     \caption{Values are grouped by part-of-speech tags based on the multiLing corpus, and are sorted in  order of difficulty. Compared to function wor
     ds, open-class words take longer to read and translate. Pearson correlations with translation difficulty (leftmost panel) are shown at the top of the predictor panels, and `*' indicates statistical significance ($p < .001$). }
\end{figure*} 

To gain more insight into the aspects of translation difficulty captured by surprisal and attention, we analyzed human difficulty and model predictions for different parts of speech. 
All results that follow are based on a subset of studies that use the same source texts, \textit{multiLing}, a small sample of news articles and sociological texts in English.\footnote{Studies included from \href{https://sites.google.com/site/centretranslationinnovation/tpr-db/public-studies}{multiLing corpus} are RUC17, ENJA15, NJ12, STC17, SG12, ENDU20 and BML12.} We break down the difficulty of these English words by their part-of-speech (POS) tags, which are available in the corpus.\footnote{POS tags are predictions of  \href{https://www.nltk.org/book/ch05.html}{NLTK tagger} converted to universal POS tags.}  

Figure~\ref{fig:src-POS} shows reading time, $s_{\mathrm{lm}}$ and attentional features of words grouped by their POS tags. Compared to function words, open-class words, such as proper nouns, nouns and adjectives, are the most difficult to translate and have higher surprisal. These words also direct more attention to \texttt{eos} and less to the context, and attract more cross-attention from the translated sequence.

For target difficulty, the distinction between open-class and function words is also evident in Figure~\ref{fig:tgt-POS}. For each source word, translation duration is defined as the duration of the target segment aligned with the source word divided by the number of alignments between the target segment and the source sentence. Translations of coordinating conjunctions and punctuation stand out as among the easiest by humans, but are surprising for the LM and difficult for NMT. 
One possible reason is that conjunctions can be cross-linguistically ambiguous \cite{li2014cross,gromann2014cross,novak2022cross}. For example, English ``but''  and ``and'' have been shown to affect NMT fluency \cite{popovic2019ambiguous,popovic2019evaluating}. For punctuation, \citet{he2019towards} demonstrate that the importance of these tokens in NMT can vary by language pairs; for example, translation to Japanese often relies on punctuation to demarcate coherent groups, which is useful for syntactic reordering.

\section{Conclusion}

Our results support the prevailing view that current NLP models, including LM and NMT align partially with human language usage and are predictive of language processing complexity. We evaluated surprisal and NMT attention as predictors of human translation difficulty, and found that both factors predict reading times and production duration. Previous work provides some evidence that surprisal and NMT attention capture important aspects of translation difficulty, and our work strengthens this conclusion by estimating surprisal based on state-of-the-art models and analyzing data based on 13 language pairs and hundreds of human translators.

Although the attentional features we consider are empirically successful and grounded in prior literature, they are not without limitations. These features are relatively simple and combining attention weights in more sophisticated ways may allow stronger predictions of human translation difficulty. 
A more theoretically-motivated approach that builds on recent studies of the interpretability of attention distributions~\cite{vashishth2019attention,zhang2021survey,madsen2022post} is  worth exploring to develop more fine-grained predictors of translation processing.

To work with as much data as possible, we focused primarily on analyses that combine data from all 13 language pairs,
but analyzing translation challenges in  individual language pairs is a high priority for future work. 
A possible next step is an analysis exploring whether the predictors considered here are sensitive to constructions in specific languages that are known sources of processing difficulty~\cite{campbell1999cognitive,vanroy2021syntactic}.
Factors such as surprisal and attentional flow are appealing in part because their generality makes them broadly applicable across languages, but understanding the idiosyncratic ways in which each pair of languages poses translation challenges is equally important.

\bibliographystyle{acl_natbib}
\bibliography{tacl}

\end{document}